\documentclass[graybox]{svmult}
\usepackage{mathptmx}       
\usepackage{helvet}         
\usepackage{courier}        
\usepackage{type1cm}        
\usepackage{makeidx}         
\usepackage{graphicx}        
\usepackage{multicol}        
\usepackage[bottom]{footmisc}
\usepackage{amsmath}
\usepackage{amssymb}
\usepackage{mathtools}

\makeindex             

\begin{document}

\title*{Optimizing Majority Voting Based Systems Under a Resource Constraint for Multiclass Problems}
\titlerunning{Optimizing Majority Voting Based Systems}
\author{Attila Tiba, Andr\'as Hajdu, Gy\"orgy Terdik and Henrietta Tom\'an}
\institute{
Attila Tiba, Andr\'as Hajdu, Gy\"orgy Terdik, Henrietta Tom\'an \at Faculty of Informatics, University of Debrecen, Hungary, \email{tiba.attila, hajdu.andras, terdik.gyorgy, toman.henrietta@inf.unideb.hu}
}
%
%
\maketitle

\abstract{Ensemble-based approaches are very effective in various fields in raising the accuracy of its individual members, when some voting rule is applied for aggregating the individual decisions. In this paper, we investigate how to find and characterize the ensembles having the highest accuracy if the total cost of the ensemble members is bounded. This question leads to Knapsack problem with non-linear and non-separable objective function in binary and multiclass classification if the majority voting is chosen for the aggregation. As the conventional solving methods cannot be applied for this task, a novel stochastic approach was introduced in the binary case where the energy function is discussed as the joint probability function of the member accuracy.
We show some theoretical results with respect to the expected ensemble accuracy and its variance in the multiclass classification problem which can help us to solve the Knapsack problem.}

\section{Introduction}
\label{sec:1}


The ensemble creation is a rather popular and effective method in several problems to outperform
the decision accuracy of individual approaches \cite{l}.
To aggregate the individual decisions of the members in the ensemble, the final decision is made by applying voting rule, such as the classic or weighted majority ones.

In a binary classification problem, each member of the ensemble makes true or false decision. It means that the classifier $D_i$ with accuracy $p_i$ ($0\leq p_i\leq 1$, $i=1,\dots,n$) can be considered as Bernoulli distributed random variable $\eta_i$, where the probability of the correct classification by $D_i$ is $p_i$. In this  particular (Bernoulli distributed) case, the expected value of the $i-$th random variable $\eta_i$ is $p_i$ $(i=1,\dots,n)$.

In majority voting, that alternative is selected as the final decision which has majority in the ensemble (more than half of the $n$ votes). In this case, the ensemble accuracy for $n\in\mathbb N$ independent binary classifiers \cite{kb} can be calculated as:
\begin{equation}
\label{eq:qcl}
q_{binary}=\sum\limits_{k=\left \lceil \frac{n}{2} \right \rceil}^n \bigg(
{\underset{|I|=k}{\sum\limits_{I\subseteq \{1,\ldots,n\}}}}
\prod\limits_{i\in I}p_i\prod\limits_{j\in \{1,\ldots,n\}\setminus
I} (1-p_j)\bigg).
\end{equation}

In \cite{Hajdu2013jspatialvoting}, the majority voting rule was extended to the spatial domain in a special object detection problem to find the optic disc (OD) in retinal images. The votes of the ensemble members (OD detectors) are given by single pixels as the centroid of the disc-like anatomical feature OD. The votes are required to fall inside a disc of a given diameter $d_{OD}$ to vote together. To aggregate the outputs of individual OD detectors, the final decision is made by choosing the circle fulfilling the geometric constraint and containing the maximal number of the votes. To find the ensemble accuracy in this case, 
the term $p_{n,k}$ is introduced for the modified majority voting of the classifiers $D_1, \ldots, D_n$: if $k$ classifiers out of the $n$ ones give a correct vote, then the good decision is made with probability $p_{n,k}$. By applying these notations, the ensemble accuracy (\ref{eq:qcl}) is transformed by the geometric restriction to the following formula:
\begin{equation}
\label{eq:spatial}
q_{multi}=\sum\limits_{k=0}^n p_{n,k}\bigg(
{\underset{|I|=k}{\sum\limits_{I\subseteq \{1,\ldots,n\}}}}
\prod\limits_{i\in I}p_i\prod\limits_{j\in \{1,\ldots,n\}\setminus
I} (1-p_j)\bigg).
\end{equation}
For the given real numbers $p_{n,k}$ $(k=0,1,\dots,n)$ in (\ref{eq:spatial}), we have that $0\leq p_{n,0}\leq p_{n,1}\leq\dots\leq p_{n,n}\leq 1$.

In special case, we get back the classical majority voting scheme if the terms $p_{n,k}$ are chosen in the following way: $p_{n,k}=1$, if $k>\left \lfloor{n/2}\right \rfloor$, and $p_{n,k}=0$, otherwise.

In the above spatial extension of the majority voting rule, the final decision is made by choosing from the candidates (circles) with respect to their cardinalities. The majority voting rule can be extended for a multiclass classification problem in a very similar way. 


High accuracy for an ensemble system is a very important and natural requirement, mainly in clinical decision making. Besides the high accuracy, other performance parameters need to be discussed, as well.  One of these parameters to be considered is the execution time. The ensemble creation is more resource demanding, because all the ensemble members have to be executed to make the final decision. 
In this paper, we solve the problem how to find the ensemble with the highest accuracy from the given possible ensemble members, with a constraint on the total execution time. These optimization problems, when the ensemble accuracy $q_{binary}$ in (\ref{eq:qcl}) or $q_{multi}$ in (\ref{eq:spatial}) 
is chosen as energy function, is very challenging,  as both of them result in a non-linear, non-separable task. It means we cannot apply the classical solving methods, namely e.g. the dynamic programming, for finding the optimal solution. 
A Knapsack problem is formulated to handle the constraint for the total execution time. We give some theoretical results with respect to the multiclass classification problem which can help us to solve the Knapsack problem.

The rest of the paper is organized as follows. In section 2, the proper formulation of the above optimization problem as Knapsack one is given.
After discussing the multiclass classification problem in contrast with the binary one in section 3, some theoretical and experimental results are enclosed for the multiclass classification problem in section 4.

\section{The Knapsack problem with total time constraint}
\label{sec:2}




As first step, the classic Knapsack problem is presented, then we formulate our ensemble creation issue and discuss why finding the solution is so difficult if the energy function of the Knapsack problem is selected as $q_{multi}$ in (\ref{eq:spatial}).




To formulate the classic Knapsack problem, let $n$ items be given, with value $v_1,\ldots, v_n$ ($v_i\geq 0$, $i=1,\dots,n$) and weight
$w_1,\ldots, w_n$ ($w_i\geq 0$, $i=1,\dots,n$), respectively. Then let $x_i$ ($x_i\in \{0,1\}$, $i=1,\dots,n$) be the number of the
$i$-th item to be packed. The maximal total weight of the knapsack is $W$ ($W\geq 0$).
The aim is to find the maximal value of the target function $\sum\limits_{k=1}^n x_k v_k$ fulfilling the following conditions:
$\sum\limits_{k=1}^n x_k w_k \leq W, \ \  x_k \in \{0,1\} \ \  (k=1,\ldots,n)$.

With respect to the corresponding properties of the objective function coming from several different kinds of applications, many variations of the original Knapsack problem are considered: linear/non-linear, separable/non-separable, 
convex/non-convex objective functions with continuous/integer variables. 
Although some non-linear Knapsack problems are investigated in the literature, 
\cite{nonlinknap1}, 
\cite{nonlinknap3}, 
the vast majority of the works deal with Knapsack problems having linear or a separable convex non-linear objective function and linear constraint. 

In the above presented ensemble creation motivated by the object detection problem, each possible ensemble member is an object detector. In Knapsack problem, the individual accuracy $p_i$ of the $i-$th detector is considered as the value $v_i$, while the individual running time $t_i$ is the weight $w_i$, where for the aggregation, a constrained majority voting is applied, that is, the ensemble accuracy $q_{multi}$ given in (\ref{eq:spatial}) is the objective function.
The problem is to find the most accurate ensemble with system accuracy $q_T$ from these members with limited total execution time $T$:

\begin{equation}
\label{qTdef} q_T= \max\limits_{\left\{i_1,\ldots ,i_s \right\}}\left\{\sum\limits_{k=0}^s p_{s,k}\bigg(
{\underset{|I|=k}{\sum\limits_{I\subseteq \{i_1,\ldots,i_s\}}}}
\prod\limits_{i\in I}p_i\prod\limits_{\mathclap{~~~~~~~~~j\in \{i_1,\ldots,i_s\}\setminus
I}} (1-p_j)\bigg)\right\}
\end{equation}
with the following conditions:\\
\begin{equation}
\label{qTcondition}
\sum\limits_{j=1}^s t_{i_j}\leq T,\ \ \left\{i_1,\ldots ,i_s\right\} \subseteq \left\{1,\ldots ,n\right\} \ \  (s=1,\ldots,n).
\end{equation}

The main challenge in solving this optimization problem is that the target function $q_{multi}$ of the constrained majority voting is non-linear, non-separable. 
In general, Knapsack problems with these special kind of objective functions are investigated very rarely in the related papers, or only in that case when a strict restriction on their functional structure is given (e.g., the exponential type of target function is analyzed in \cite{nonlinknap3}). 
That is, for a proper analysis we need some theoretical results for the optimization of the specific target function (\ref{eq:spatial}) within the Knapsack framework.




\section{The multiclass classification problem}
\label{sec:3}

In binary classification, the elements of a given set are classified into two classes (predicting which class each element belongs to). 
As first step, a Knapsack problem is investigated for ensemble creation with binary classifiers $D_1, D_2, \dots, D_n$ as possible members of the ensemble, whose outputs are aggregated by applying the majority voting rule. It means that in this Knapsack problem, the objective function $q_{binary}$ given in (\ref{eq:qcl}) is maximized when the total execution time of the selected members is bounded (see the condition in (\ref{qTcondition})).

In our proposed stochastic approach in \cite{7899637}, the selection of the items to the ensemble is based on the efficiency of the individual members. Instead of the usefulness values $p_{i}/t_{i}$ considered in the classic greedy method, the system accuracy $q(p_{i},t_{i})$ of the ensemble containing maximal number 
of $i$-th items 
characterizes the efficiency of the $i$-th kind of item.

In our selection method, a discrete random variable depending on the efficiency values of the remaining items is applied in each step to determine the probability of choosing an item from the remaining set to add to the ensemble. This discrete random variable reflects that the more efficient the item is, the more probable it is selected to the ensemble in the next step.

To find and apply proper stopping criteria for this selection method, the behavior of the random variable $q_{binary}$, the joint distribution function based on the values $p_i$-s in (\ref{eq:qcl}) is investigated. Either the distribution of the values $p_i$ is known, or it is fitted by Beta distribution, the knowledge on the behavior of the energy function $q_{binary}$ (e.g. the expected ensemble accuracy, the probability to find more accurate ensembles) can be efficiently involved as a stopping rule in the stochastic search.

The multiclass classification can be interpreted in a similar way as the binary one, just in case the prediction of the class for each element where it belongs to is made for three or more classes \cite{multiclass}.
We encounter similar problems to find the optimal solution $q_T$ in (\ref{qTdef}) of multiclass Knapsack problem as in the binary case, but, besides the estimation of the behavior of the energy function $q_{multi}$, the terms $p_{n,k}$ need to be investigated, as well. It is reasonable to assume that the more classifiers out of the $n$ ones give correct vote, the bigger probability $p_{n,k}$ for the good decision we get for the ensemble. Therefore, in the next section, the terms $p_{n,k}$ are considered as values of a function $F$ such that $p_{n,k}= F\left(\frac{k}{n}\right)$, where $F\left(\cdot\right)  $ is a cumulative distribution function on $\left[  0,1\right]$.

\section{Stochastic estimation of ensemble accuracy}
\label{sec:4}
We have the following theorem showing the behavior of the random variable $q_{multi}$ (i.e. the expected ensemble accuracy and the variance), based on the random values of $p_i$-s.

\begin{theorem}
\label{Bern}
Let $p\in\left[  0,1\right]$ be a random variable with $Ep=\mu$, Var$\left(
p\right)  =\sigma^{2}$, and $p_{i}$ $(i=1,2,\ldots,n)$ are independent and identically distributed according to $p$. Furthermore let the energy function $q_{multi}$ be defined by (\ref{eq:spatial}).
Then for the expected ensemble accuracy $E(q_{multi})$ we have shown that
\begin{equation}
E(q_{multi})=\sum_{k=0}^{n}F\left(  \frac{k}{n}\right)  \binom{n}{k}\mu^{k}\left(
1-\mu\right)  ^{n-k}.%
\end{equation}
Furthermore, if $n$ is large then
\begin{equation}
\sum_{k=0}^{n}F\left(  \frac{k}{n}\right)  \binom{n}{k}\mu^{k}\left(
1-\mu\right)  ^{n-k} \sim\int_{0}^{1}F\left(  y\right)  \delta\left(  \mu\right)  dy=F\left(
\mu\right)
\end{equation}
where $ \delta\left( \cdot \right)$ is the Dirac function.

In case of large $n$,  we have  the variance of the ensemble accuracy
\begin{equation}
0\leq \operatorname*{Var}\left(  q_{multi}\right)\leq F\left(  \mu\right)  -F^{2}\left(  \mu\right)  =F\left(  \mu\right)
\left(  1-F\left(  \mu\right)  \right).
\end{equation}
\end{theorem}


For practical issue, the following examples for the function $F$ are important:

Arcsine law (distributed as Beta $\left(  1/2,1/2\right)  $) with
cumulative distribution function
\begin{equation}
F\left(  y\right)  =\frac{2}{\pi}\arcsin\left(  \sqrt{y}\right)  ,\quad
y\in\left[  0,1\right],
\end{equation}
and Generalized Arcsine law (distributed as Beta $\left(  1-\alpha,\alpha\right)$), as if the distribution of $p$ is not known, then a Beta distribution is fitted to $p$.

From the results of the Theorem \ref{Bern} with respect to the expected value and the variance of the ensemble accuracy, the decision in the multiclass case for relatively large $n$ is considered to be Bernoulli variated with parameter $F\left(  \mu\right)$.

While the binary classification problem is closely related to the results of the binomial distribution, then in the multiclass classification the multinomial coefficients are supposed to have very important role in finding a formula for the values of $p_{n,k}(d)$. 
As a first step, we simulated the multiclass classification problem for $d=3$, $d=4$ and $d=5$
classes, 
by generating random numbers in $\left[  0,1\right]$, to decide which class is chosen. From the results of the simulations, we get approximate values for the terms $p_{n,k}(d)$. 
In the next step, we give a closed formula for the values $p_{n,k}(d)$, as well. 

Let the multinomial coefficients $b_{n,d}\left(  x_{1},x_{2},\ldots, x_{d}\right)$ be given, ($x_{i}\geq0$, $\sum x_{i}=n$), $\underline{x}=\left(  x_{1},x_{2},\ldots, x_{d}\right)$, and $\alpha_{k}\left(  \underline{x}\right)  $ is defined as the $\operatorname{card}%
\left(  \left.  \underline{x}\right\vert x_{i}=k\right)  +1$.
Then for the terms  $\mathfrak{p}_{n,k}(d)$ of accuracy in that case, we have the following formula,
\begin{equation}
{\mathfrak{p}}_{n,k}\left(  d\right)  =\frac
{1}{d^{n-k}}\sum_{0\leq\underline{x}\leq k}\frac{b_{n-k,d}%
\left(  \underline{x}\right)  }{\alpha_{k}\left(  \underline{x}\right)  },%
\end{equation}
where $0\leq\underline{x}\leq k:= (x_i| 0 \leq x_i \leq k,  i=1,2,\ldots,d)$.

Applying this formula, we get the same results for the values of $\mathfrak{p}_{n,k}(d)$ in case of $d=3$, $d=4$ and $d=5$ classes as before with the simulations.

The closed formula for the values of $\mathfrak{p}_{n,k}(d)$  guarantee us that besides the experimental  results (e.g. simulations), further theoretical investigation and characterization of the optimal solution of the Knapsack problem in multiclass classification can be achieved as our future plan. 



\begin{acknowledgement}
This work is supported in part by the project EFOP-3.6.2-16-2017-00015 supported by the European Union, co-financed by the European Social Fund.

\end{acknowledgement}

%
%
%
\biblstarthook{
}

\end{document}